\documentclass[10pt,twocolumn,letterpaper]{article}

\usepackage[pagenumbers]{cvpr} 
\definecolor{cvprblue}{rgb}{0.21,0.49,0.74}
\usepackage[pagebackref,breaklinks,colorlinks,allcolors=cvprblue]{hyperref}

\usepackage{multirow}
\usepackage{bm}

\def\paperID{24} 
\def\confName{CVPR}
\def\confYear{2026}

\title{Benchmarking Vision Foundation Models\\for Domain-Generalizable Face Anti-Spoofing}

\author{Mika Feng, Pierre Gallin-Martel, Koichi Ito, and Takafumi Aoki\\
Graduate School of Information Sciences, Tohoku University, Japan\\
{\tt\small \{mika, pierre, ito\}@aoki.ecei.tohoku.ac.jp, aoki@ecei.tohoku.ac.jp}
}

\begin{document}
\maketitle
\begin{abstract}
Face Anti-Spoofing (FAS) remains challenging due to the requirement for robust domain generalization across unseen environments. While recent trends leverage Vision-Language Models (VLMs) for semantic supervision, these multimodal approaches often demand prohibitive computational resources and exhibit high inference latency. Furthermore, their efficacy is inherently limited by the quality of the underlying visual features. This paper revisits the potential of vision-only foundation models to establish a highly efficient and robust baseline for FAS. We conduct a systematic benchmarking of 15 pre-trained models, such as supervised CNNs, supervised ViTs, and self-supervised ViTs, under severe cross-domain scenarios including the MICO and Limited Source Domains (LSD) protocols. Our comprehensive analysis reveals that self-supervised vision models, particularly DINOv2 with Registers, significantly suppress attention artifacts and capture critical, fine-grained spoofing cues. Combined with Face Anti-Spoofing Data Augmentation (FAS-Aug), Patch-wise Data Augmentation (PDA) and Attention-weighted Patch Loss (APL), our proposed vision-only baseline achieves state-of-the-art performance in the MICO protocol. This baseline outperforms existing methods under the data-constrained LSD protocol while maintaining superior computational efficiency. This work provides a definitive vision-only baseline for FAS, demonstrating that optimized self-supervised vision transformers can serve as a backbone for both vision-only and future multimodal FAS systems.

\end{abstract}    
\section{Introduction}
\label{sec:intro}

Face recognition technology, which identifies individuals based on unique facial biometrics, has become a cornerstone of modern authentication systems due to its inherent advantages, including low implementation cost, contactless operation, and high user convenience \cite{Handbook-Face-Recognition}.
While face recognition systems are engineered for robustness against common environmental variations, such as changes in head pose, illumination, and image blur, this resilience is paradoxically exploited by malicious actors through spoofing attacks.
Specifically, the ease with which presentation materials (e.g., printed photos or replayed videos) can be used to impersonate a registered user poses a critical security threat to face recognition systems. 
These attempts are known as Presentation Attacks (PAs), and they are becoming increasingly prevalent since high-quality face images are accessible online, making PAs a realistic and easily executed method to bypass security protocols \cite{Handbook-Anti-Spoofing}.

To effectively counter PAs, the core challenge for Face Anti-Spoofing (FAS) is to detect minute and fine-grained discrepancies between a genuine live face and a spoofed presentation.
This requires extracting subtle but highly discriminative features that reveal the artifacts of the presentation medium, such as the distinctive texture of paper, the moir\'{e} patterns from a display device, or the lack of authentic 3D information such as depth and light reflection.
Consequently, numerous deep learning methods utilizing Convolutional Neural Networks (CNNs) and Vision Transformers (ViT) \cite{Dosovitskiy-ICLR-2021} have been proposed to learn these inherent spoofing cues \cite{Yu-CVPR-2020,Yu-PAMI-2020,Wang-CVPR-2022,Watanabe-APSIPA-2022,Chen-CoRR-2023,Wang-TBIOM-2022,Li-NN-2024,Zheng-IFS-2024,He-CVPRW-2024,Feng-CVPRW-2025,Feng-ICCVW-2025}.
While these methods often achieve high accuracy in controlled settings (i.e., intra-dataset evaluation), their primary limitation emerges when applied to real-world scenarios.
Their performance degrades significantly when encountering unseen attack types, new acquisition devices, or novel environmental conditions.
This is known as domain gap in FAS. 

This domain gap, which leads to degraded detection accuracy against unseen attacks, primarily stems from two technical factors.
The first is the limited capability of the model structure to resist domain-specific biases acquired during training, and the second is the deficiency of the feature extractor in producing robust domain-invariant features across diverse real-world scenarios.
So far, CNN-based FAS methods, such as CDCN \cite{Yu-CVPR-2020}, NAS-FAS \cite{Yu-PAMI-2020}, and PatchNet \cite{Wang-CVPR-2022}, have focused on extracting fine-grained local features inherent to spoofing attacks.
Notably, PatchNet reformulated FAS as a fine-grained patch-type recognition problem and improved generalization by learning highly discriminative features from local capture characteristics.
However, many of these CNN-based methods suffer from an inherent lack of generalization capability in their initial parameters against unknown attacks and environmental changes, primarily because they do not leverage large-scale pre-trained models as their backbones.
Recently, FAS methods utilizing ViT \cite{Watanabe-APSIPA-2022,Wang-TBIOM-2022,Chen-CoRR-2023,Feng-CVPRW-2025} have been proposed, demonstrating superiority over CNNs in their ability to capture both local and global features integrally.
Nevertheless, most pre-trained ViTs employed by these methods rely on supervised learning with limited image datasets, resulting in limitations in their capacity to extract robust and domain-invariant generic features.
For instance, while Segment Anything Model (SAM) \cite{Kirillov-ICCV-2023} used by Chen et al. \cite{Chen-CoRR-2023} is ViT-based, it is specifically designed for segmentation, not for extracting the discriminative liveness cues required by the FAS task.
Therefore, we consider that leveraging a vision foundation model, which is pre-trained on massive amounts of data in a self-supervised manner and possesses high versatility and robustness, is crucial for the feature extractor to achieve domain generalization in FAS.

Recently, several approaches have emerged that leverage Vision-Language Models (VLMs) to enhance generalization through semantically rich textual supervision \cite{Srivatsan-ICCV-2023,Liu-ICCV-2023,Zhang-AAAI-2025}.
However, these VLM-based multimodal techniques typically require massive computational resources (e.g., large-scale LLMs) and suffer from slow inference speeds, making them difficult to deploy in real-world, resource-constrained authentication systems.
Furthermore, their performance remains inherently constrained by the quality and domain-invariance of the underlying visual features.
Therefore, instead of simply adopting computationally expensive VLMs, achieving highly efficient and robust domain generalization in FAS requires a systematic evaluation of vision-only foundation models to establish a solid baseline.

To explore the potential of vision-only models, Feng et al. demonstrated the utility of intermediate ViT features for FAS \cite{Feng-CVPRW-2025}.
They also demonstrated that DINOv2 with Registers \cite{Darcet-ICLR-2024} effectively suppresses attention artifacts and captures fine-grained spoofing cues \cite{Feng-ICCVW-2025}.
However, their evaluations were limited to intra-dataset protocols (e.g., SiW \cite{Liu-CVPR-2018}, OULU-NPU \cite{Boulkenafet-FG-2017}), which fail to measure the robustness against real-world domain shifts.
In this paper, we extend this research direction by conducting a comprehensive analysis of various vision foundation models specifically under severe cross-domain scenarios (e.g., the MICO protocol and Limited Source Domains).
Specifically, we comprehensively evaluate 15 different visual feature extractors, encompassing supervised CNNs, supervised ViTs, and self-supervised ViTs.
Through our extensive benchmarking, we demonstrate that DINOv2 with Registers \cite{Darcet-ICLR-2024}, combined with Face Anti-Spoofing Data Augmentation (FAS-Aug) \cite{Cai-IJCV-2024}, Patch-wise Data Augmentation (PDA) \cite{Watanabe-APSIPA-2022}, and Attention-weighted Patch Loss (APL) \cite{Watanabe-APSIPA-2022} serves as the most effective and robust feature extractor.
By establishing this strong vision-only baseline, we show that an optimized self-supervised vision transformer can achieve highly competitive domain generalization comparable to existing methods in the MICO protocol.
Furthermore, our baseline achieves higher accuracy than existing methods under the Limited Source Domains (LSD) protocol, demonstrating the effectiveness of DINOv2 in data-constrained scenarios.
Consequently, our comprehensive analysis provides a practical and generalizable vision-only baseline for FAS, offering an effective visual feature extractor that can serve as a strong backbone for both vision-only methods and advanced multimodal approaches.

\begin{figure*}[t]
  \centering
  \includegraphics[width=\linewidth]{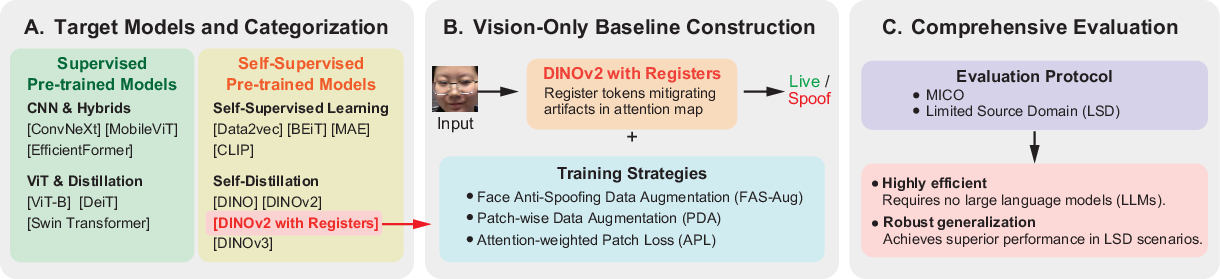}
  \caption{Overview of our comprehensive benchmarking framework and the proposed vision-only baseline. 
(A) Target Models and Categorization: We systematically categorize and evaluate 15 diverse vision foundation models across different pre-training paradigms (Supervised CNNs, Supervised ViTs, and Self-Supervised \& Multi-modal ViTs). 
(B) Vision-Only Baseline Construction: Based on the benchmark insights, we construct a highly efficient baseline utilizing DINOv2 with Registers to mitigate attention artifacts, further enhanced by robust training strategies including FAS-Aug, PDA, and APL. 
(C) Comprehensive Evaluation: Extensive evaluations under the standard MICO and Limited Source Domain (LSD) protocols demonstrate that our vision-only approach achieves superior domain generalization.}
  \label{fig:benchmark}
\end{figure*}

\section{Related Work}

This section gives an overview of research in FAS, ranging from traditional to advanced foundation model-based approaches.

\subsection{Early FAS Methods}

FAS is essential for securing face recognition systems against PAs such as printed photos and 3D masks \cite{Handbook-Anti-Spoofing}.
While early methods relied on handcrafted features \cite{Scenario?,IDIAP,Context-based,Unlock,Speeded-UP}, the field quickly shifted to deep learning.
CNN-based approaches, such as PatchNet \cite{Wang-CVPR-2022} and NAS-FAS \cite{Yu-PAMI-2020}, achieve high accuracy in controlled, intra-domain settings by learning discriminative local representations or utilizing auxiliary cues \cite{Liu-CVPR-2018,Wang-CoRR-2018,Yu-CVPR-2020}.
However, these methods often suffer from severe performance degradation when exposed to unseen datasets or novel attack types.

\subsection{Domain Generalization for FAS}

To address the cross-domain challenge, various Domain Generalization (DG) strategies have been proposed to learn domain-invariant representations.
These methods employ meta-learning (e.g., RFMeta \cite{RFMeta}, MADDG \cite{Shao-CVPR-2019}), feature disentanglement (e.g., DR-MD-Net \cite{Wang-CVPR-2020}, MFAE \cite{Zheng-IFS-2024}), or domain alignment techniques \cite{Chen-AAAI-2021,Liu-ECCV-2022,Liu-IJCAI-2021,Liu-ACM-2021,SSAN,Liao-WACV-2023,SA-FAS,Zhou-CVPR-2023,Le-CVPR-2024}.
Recent approaches also leverage physics-based data synthesis \cite{Cai-IJCV-2024} to simulate diverse attack variations.
Despite these advancements, achieving true domain invariance against real-world shifts (e.g., unseen capture devices, lighting conditions, and novel spoof materials) remains a highly challenging and open problem.

\subsection{Vision Foundation Models in FAS}

Recently, the pursuit of robust representations has driven FAS research toward large-scale foundation models.
Several approaches leverage Vision-Language Models (VLMs), such as CLIP \cite{Radford-ICML-2021}, or employ multimodal architectures (e.g., FLIP \cite{Srivatsan-ICCV-2023}, I-FAS \cite{Zhang-AAAI-2025}, InstructFLIP \cite{Lin-ACM-2025}) to enhance generalization through semantic text guidance.
However, these VLM-based methods typically demand massive computational resources and suffer from high inference latency, limiting their practical deployment in real-world authentication systems.
In contrast, self-supervised vision-only foundation models like DINO \cite{Caron-ICCV-2021} and DINOv2 \cite{Oquab-TMLR-2024} provide a highly efficient and robust visual foundation.
They are effective at capturing fine-grained, localized spatial features essential for detecting minute spoofing artifacts without relying on semantic labels.
Specifically, DINOv2 with Registers \cite{Darcet-ICLR-2024} introduces auxiliary tokens to suppress attention artifacts.
This stabilization has proven crucial for FAS, allowing the model to accurately capture fine-grained spoofing cues \cite{Feng-ICCVW-2025}.
While these vision models show significant promise, a comprehensive benchmarking of their domain generalization capabilities in FAS is still lacking.
In this study, we revisit vision-only foundation models for FAS and establish a strong baseline through comprehensive benchmarking.

\begin{figure*}[t]
  \centering
  \includegraphics[width=\linewidth]{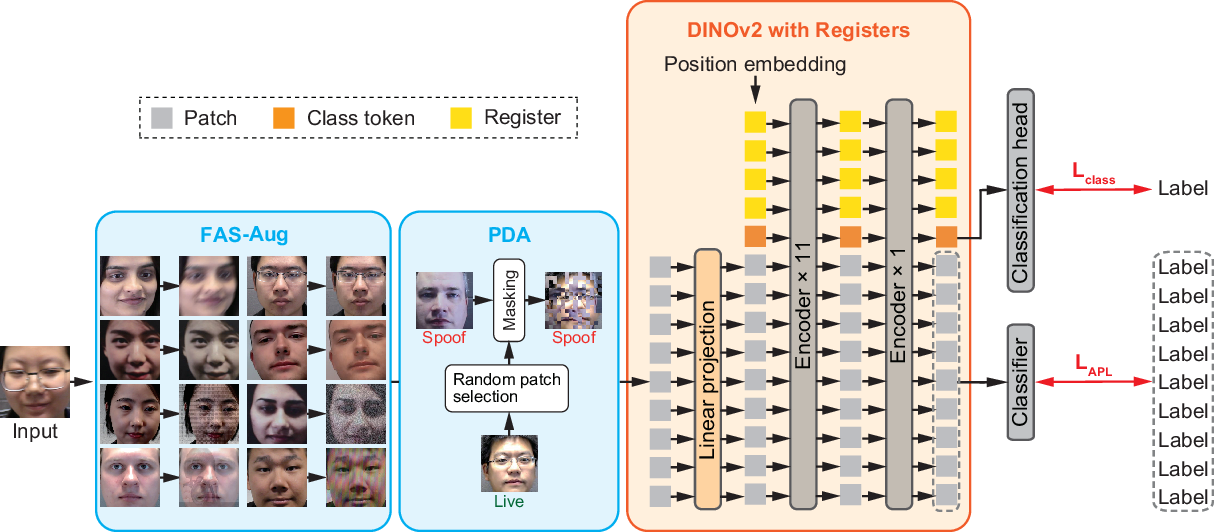}
  \caption{Overview of our proposed vision-only baseline.
  During training, input images are augmented by Face Anti-Spoofing Data Augmentation (FAS-Aug) and Patch-wise Data Augmentation (PDA) to simulate diverse spoofing artifacts and enforce robust local representations.
  The processed images are then fed into the DINOv2 with Registers backbone.
  Finally, the network is optimized using a dual-level supervision strategy: a global classification loss ($\mathcal{L}_{class}$) derived from the class token, and a local Attention-Weighted Patch Loss ($\mathcal{L}_{APL}$) applied to the patch tokens.}
  \label{fig:baseline}
\end{figure*}
\section{Benchmarking Framework and Baseline}

The primary goal of this section is to establish a systematic benchmarking framework to identify the most effective vision foundation models for FAS.
We first categorize a set of pre-trained models used for feature extraction.
Subsequently, we describe the construction of our baseline model, which integrates the most promising backbone with FAS-Aug \cite{Cai-IJCV-2024}, PDA \cite{Watanabe-APSIPA-2022} and APL \cite{Watanabe-APSIPA-2022}.

\subsection{Categorization of Vision Foundation Models}
\label{sec:vfm}

To systematically evaluate how different architecture designs and pre-training strategies contribute to robustness against spoofing attacks, we classify the selected models along two primary axes: architecture designs and pre-training strategies.

\subsubsection{Architecture Design}

We evaluate models across a spectrum of architectures, ranging from modernized convolutions to attention mechanisms, to understand their baseline feature extraction capabilities.

\noindent
\textbf{CNNs and Hybrids}: 
ConvNeXt \cite{Liu-CVPR-2022} serves as a modernized CNN designed to achieve performance competitive with Vision Transformers (ViTs) by incorporating recent architectural advancements.
To address the trade-off between representational power and efficiency, we include MobileViT \cite{Mehta-ICLR-2022}, a lightweight hybrid model that integrates the inductive biases of CNNs with the global modeling capabilities of Transformers, and EfficientFormer \cite{Li-NeurIPS-2022}, which optimizes Transformer structures for high-speed inference.

\noindent
\textbf{Vision Transformers}: 
We evaluate ViT-B \cite{Dosovitskiy-ICLR-2021}, the standard ViT backbone that captures global context by processing images as a sequence of patches. Additionally, we include the Swin Transformer \cite{Liu-ICCV-2021}, which achieves multi-scale representations through a hierarchical structure and shifted window-based self-attention.

\subsubsection{Pre-training Strategies}

Beyond network architecture, the method of pre-training significantly impacts a model's ability to extract robust, domain-invariant features critical for cross-domain FAS.

\noindent
\textbf{Supervised Learning and Distillation}: 
Several models in our benchmark rely on traditional supervised learning.
To improve data efficiency in this setting, we evaluate DeiT \cite{Touvron-PMLR-2021}, which utilizes a teacher-student distillation strategy to train ViTs effectively without relying on massive, fully-annotated datasets.

\noindent
\textbf{Self-Supervised Learning (SSL)}: 
SSL models have demonstrated exceptional transferability across diverse downstream tasks by learning without explicit labels.
We include Masked Image Modeling (MIM) approaches such as BEiT \cite{Bao-ICLR-2022} and MAE \cite{He-CVPR-2022}.
While BEiT treats images as discrete tokens to learn semantic representations by predicting masked patches, MAE focuses on reconstructing original pixel values from highly masked inputs.
We also evaluate Data2vec \cite{Baevski-PMLR-2022}, which learns the essential structure of input data by predicting latent representations of augmented views, and CLIP \cite{Radford-ICML-2021}, a vision-language foundation model that gains semantically aligned visual representations through contrastive learning on massive image-text pairs.

\noindent
\textbf{Self-Distillation (DINO Family)}: 
Central to our analysis is the DINO family, which relies on a self-distillation paradigm. Starting from the original DINO \cite{Caron-ICCV-2021}, we evaluate DINOv2 \cite{Oquab-TMLR-2024} and its refined variant, DINOv2 with Registers \cite{Darcet-ICLR-2024}.
The latter specifically addresses feature map artifacts by introducing dedicated register tokens, yielding smoother representations that are critical for capturing fine-grained spoofing cues.
Finally, we include DINOv3 \cite{Siméoni-arXiv-2025}, which scales the training strategies of the DINOv2 framework to further enhance adaptability to complex visual tasks.

\subsection{Baseline Method}

This section describes the proposed baseline method, designed to enhance FAS generalization against unseen attacks.
Recognizing that the rich visual representations learned through massive-scale self-supervised pre-training significantly improve detection performance, we adopt the vision foundation model DINOv2 \cite{Oquab-TMLR-2024} as our core architecture.
Specifically, we utilize DINOv2 with Registers \cite{Darcet-ICLR-2024} to mitigate attention perturbations and focus on fine-grained visual cues.
Moreover, robustness across diverse spoofing scenarios is achieved by combining FAS-Aug \cite{Cai-IJCV-2024} and PDA \cite{Watanabe-APSIPA-2022} with a dual-level loss function incorporating Attention-Weighted Patch Loss (APL) \cite{Watanabe-APSIPA-2022}.
The overall pipeline is illustrated in Fig. \ref{fig:baseline}.

\subsubsection{Model Architecture}
\label{sec:architecture}

The baseline model initiates by fine-tuning the DINOv2 ViT-B/14 with Registers \cite{Darcet-ICLR-2024} without freezing any layers.
A $224 \times 224$ pixel input image is divided into $14 \times 14$ non-overlapping patches and processed through the 12-layer Transformer encoder.
The incorporation of register tokens, defined as learnable tokens appended to the input sequence, is critical for achieving the fine-grained feature extraction required for FAS.
Prior studies demonstrate that standard ViTs suffer from an ``attention spike'' phenomenon, where anomalously high attention concentrates on irrelevant background areas due to the model repurposing patch tokens for global information storage \cite{Darcet-ICLR-2024}.
Register tokens mitigate this noise, effectively stabilizing the attention mechanism and yielding smoother, more interpretable attention maps.
This enables the model to accurately capture the minute visual cues essential for spoof detection.
A classification head is appended to the final layer.
During inference, the output probability of the class token from this head, $P_{Live}$, is computed.
The image is classified as ``Live'' if $P_{Live}$ exceeds a predetermined threshold, and ``Spoof'' otherwise.
This threshold is determined as the Equal Error Rate (EER) point on the testing set, where the False Acceptance Rate (FAR) equals the False Rejection Rate (FRR) \cite{Watanabe-APSIPA-2022}.

\subsubsection{Data Augmentation}

To maximize the model's robustness and generalization capability across domains, we integrate two complementary augmentation techniques applied exclusively during the training phase.
First, FAS-Aug \cite{Cai-IJCV-2024} is employed to simulate various forms of degradation and attack artifacts.
This strategy encompasses eight augmentation types, including photography noise (e.g., hand trembling, low resolution), print attack artifacts (e.g., color distortion), and display attack artifacts (e.g., moir\'{e} patterns).
Photography noise is applied without altering the label, while simulations of print and display attacks result in the label being reassigned to ``Spoof.''
During training, one of these eight augmentations is randomly applied to each input image based on its original specification.
Second, we utilize PDA \cite{Watanabe-APSIPA-2022}, a patch-level augmentation method tailored for Vision Transformers.
Specifically, we employ the Live Patch Mask component, which randomly replaces certain patches in a spoofed image with corresponding patches from a live image with a probability of $P=0.5$.
The model is then trained to classify the substituted live patches as ``Live,'' the remaining spoofed patches as ``Spoof,'' and the overall image as ``Spoof.''
This patch-level mixing increases task difficulty, compelling the model to learn highly discriminative and robust local representations.

\subsubsection{Loss Function}

During network training, the baseline method employs a dual-level loss function
to facilitate robust learning at both the global (image) and local (patch) levels:\begin{equation}
  \mathcal{L}_{total} = \mathcal{L}_{class}+\mathcal{L}_{APL}.\end{equation}
The global loss, $\mathcal{L}_{class}$, corresponds to the standard binary classification loss computed from the class token output.
The local loss, APL \cite{Watanabe-APSIPA-2022} ($\mathcal{L}_{APL}$), is computed for each patch using the attention map of the class token from the 12th encoder block as a weighting factor.
This patch-wise supervision enhances the model's ability to detect localized spoofing artifacts.
Furthermore, for the binary classification within $\mathcal{L}_{APL}$, the L2-constrained Softmax loss \cite{Ranjan-CoRR-2017} is adopted to ensure balanced feature learning between the ``Live'' and ``Spoof'' classes, thereby improving classification stability.
The L2-constrained Softmax loss enforces all feature vectors to have a fixed L2 norm $\alpha$ and is defined by:
\begin{equation}
  {\rm L2Softmax} = -\frac{1}{N} \sum_{i=1}^N \log \frac{e^{W_{y_i}^{\top} f(\bm{x}_i)+b_{y_i}}}{\sum_{j=1}^C e^{W_j^{\top} f(\bm{x}_i)+b_j}},\end{equation}
\begin{equation*}
  {\rm s.t.} \ \ \| f(\bm{x}_i) \|_2 = \alpha, \quad \forall i = 1, 2, \ldots, N,
\end{equation*}
where $\bm{x}_i$ is the input image, $N$ is the mini-batch size, $W$ is the weight matrix of the fully-connected layer, $f(\bm{x}_i)$ is the extracted feature vector, $C$ is the number of classes, $\alpha$ is a hyperparameter, $W_{y_i}$ is the column of $W$ corresponding to the ground-truth label $y_i$, and $b_{y_i}$ is the bias term.

\section{Experiments and Discussion}

This section describes experiments to demonstrate the effectiveness of the baseline method in detecting face spoofing attacks.


\begin{table*}[t]
  \centering
  \caption{Comparison of various vision foundation models as feature extractors for FAS. Evaluation metrics are reported in AUC [\%]$\uparrow$. Models are categorized into: \colorbox{blue!15}{Supervised CNNs}, \colorbox{green!15}{Supervised ViTs}, and \colorbox{yellow!40}{Self-Supervised \& Multi-modal ViTs}. The best and second-best results are highlighted in \textbf{bold} and \underline{underline}, respectively.}
  \label{tbl:exp_1}
  \begin{tabular}{lcccccc}
    \hline
    Model 
    & CIO$\rightarrow$M $\uparrow$ 
    & OMI$\rightarrow$C $\uparrow$ 
    & OCM$\rightarrow$I $\uparrow$ 
    & ICM$\rightarrow$O $\uparrow$
    & Avg. $\uparrow$\\
    \hline
    \colorbox{blue!15}{ConvNeXt \cite{Liu-CVPR-2022}} 
    & 95.95 
    & 95.78 
    & 84.87 
    & 84.26 
    & 90.22 \\
    \hline
    \colorbox{green!15}{ViT-B \cite{Dosovitskiy-ICLR-2021}}
    & 91.81 
    & 96.82 
    & \bf{95.30} 
    & 91.96
    & 93.97 \\
    \colorbox{green!15}{EfficientFormer \cite{Li-NeurIPS-2022}}
    & 93.67 
    & 93.55 
    & 81.61 
    & 84.55
    & 88.35 \\
    \colorbox{green!15}{MobileViT \cite{Mehta-ICLR-2022}}
    & 96.73 
    & 94.99 
    & 82.92 
    & 90.80
    & 91.36 \\
    \colorbox{green!15}{DeiT-tiny \cite{Touvron-PMLR-2021}}
    & 90.10 
    & 69.40 
    & 68.09 
    & 78.63
    & 76.56 \\
    \colorbox{green!15}{DeiT-base \cite{Touvron-PMLR-2021}}
    & 88.82 
    & 83.21 
    & 77.36 
    & 75.17
    & 81.14 \\
    \colorbox{green!15}{Swin Transformer \cite{Liu-ICCV-2021}}
    & 97.20 
    & \underline{98.78} 
    & 92.44 
    & 93.42
    & 95.46 \\
    \hline
    \colorbox{yellow!40}{Data2vec \cite{Baevski-PMLR-2022}}
    & 88.82 
    & 95.13 
    & 77.36 
    & 93.47
    & 88.70 \\
    \colorbox{yellow!40}{BEiT \cite{Bao-ICLR-2022}}
    & 97.25 
    & 95.80 
    & 82.24 
    & 88.89
    & 91.05 \\
    \colorbox{yellow!40}{MAE \cite{He-CVPR-2022}}
    & 94.41 
    & 90.00 
    & 84.06 
    & 91.84
    & 90.08 \\
    \colorbox{yellow!40}{CLIP \cite{Radford-ICML-2021}}
    & \bf{98.52} 
    & 94.95 
    & 92.20 
    & 95.95
    & 95.41\\
    \colorbox{yellow!40}{DINO \cite{Caron-ICCV-2021}}
    & 97.01 
    & 97.00 
    & 91.12 
    & 94.80 
    & 94.98 \\
    \colorbox{yellow!40}{DINOv2 \cite{Oquab-TMLR-2024}}
    & 96.90 
    & 98.02 
    & 91.05 
    & \underline{96.23}
    & \underline{95.55} \\
    \colorbox{yellow!40}{DINOv2 with Registers \cite{Darcet-ICLR-2024}}
    & \underline{97.47} 
    & 98.30 
    & \underline{93.83} 
    & 95.38
    & \textbf{96.25} \\
    \colorbox{yellow!40}{DINOv3 \cite{Siméoni-arXiv-2025}}
    & 95.42 
    & \bf{98.80} 
    & 89.21 
    & \bf{97.13}
    & 95.14\\
    \hline
  \end{tabular}
\end{table*}

\subsection{Evaluation Protocols and Metrics}

We conduct cross-dataset evaluations to measure the generalization capability of the FAS methods to unseen domains.
We follow the widely adopted MICO protocol involving four benchmark datasets: MSU MFSD \cite{MSU-MFSD} (M), IDIAP Replay Attack \cite{IDIAP} (I), CASIA-FASD \cite{CASIA-FASD} (C), and OULU-NPU \cite{Boulkenafet-FG-2017} (O).
This protocol employs a leave-one-out strategy, where the model is trained on three datasets and tested on the remaining unseen dataset.
To further evaluate robustness under data-constrained conditions, we also conduct a Limited Source Domain (LSD) evaluation, using only two datasets for training.
Performance is evaluated by the Half Total Error Rate (HTER) and the Area Under the ROC Curve (AUC).


\subsection{Datasets and Preprocessing}

The datasets used for evaluation are 
the four aforementioned datasets for cross-dataset testing.
All video frames are preprocessed to extract the face region, as the input requires single images focused on the face.
For all four datasets used in the cross-dataset evaluation, Dlib \cite{Dlib} is utilized for face detection.
All extracted face regions are resized to $224 \times 224$ pixels, and pixel values are normalized to have zero mean and unit variance per channel.
The FAS methods used in the experiments method assumes a single image input rather than a video sequence. 
Therefore, frames are sampled from the videos.
For the cross-dataset evaluation, a total of 5 frames are sampled at a fixed interval.

\subsection{Implementation Details}

The proposed baseline is fine-tuned using the AdamW \cite{AdamW} optimizer, with training running for 200 epochs and utilizing early stopping based on a patience of 20 epochs.
The mini-batch size is set to 32.
In the proposed baseline, DINOv2 ViT-B/14 with Registers \cite{Darcet-ICLR-2024} pre-trained model is fine-tuned using the training data specified by the evaluation protocol.
Training inputs are augmented with FAS-Aug \cite{Cai-IJCV-2024} ($P=0.2$), PDA \cite{Watanabe-APSIPA-2022} ($P=0.2$), and standard augmentations (Random Horizontal Flip, Random Rotation, and Random Brightness). 
The learning rates are set as $5 \times 10^{-5}$ for the classification head and $5 \times 10^{-6}$ for the DINOv2 encoder.


\subsection{Experiments and Discussion}
\label{sec:results}

We conduct four sets of experiments to validate our baseline: (i) Comparison of vision foundation models to identify the most effective feature extractor; (ii) Cross-dataset evaluation under the MICO protocol; (iii) LSD evaluation; and (iv) Analysis of computational efficiency.

\noindent
{\bf (i) Comparison of Vision Foundation Models}:
In this experiment, we evaluate the effectiveness of various pre-trained backbones to determine the optimal feature extractor for FAS.
To ensure a fair comparison of the inherent representation power of backbones, FAS-Aug, PDA and $\mathcal{L}_{APL}$ were not applied in this specific set of experiments.
We compare models categorized by their pre-training paradigms as defined in Sect. \ref{sec:vfm}:
\begin{itemize}
  \item Supervised Models: ConvNeXt \cite{Liu-CVPR-2022} (CNN), ViT-B \cite{Dosovitskiy-ICLR-2021}, EfficientFormer \cite{Li-NeurIPS-2022}, MobileViT \cite{Mehta-ICLR-2022}, DeiT \cite{Touvron-PMLR-2021}, and Swin Transformer \cite{Liu-ICCV-2021}.
  \item Self-Supervised \& Multi-modal Models: Data2vec \cite{Baevski-PMLR-2022}, BEiT \cite{Bao-ICLR-2022}, MAE \cite{He-CVPR-2022}, CLIP \cite{Radford-ICML-2021}, DINO \cite{Caron-ICCV-2021}, DINOv2 \cite{Oquab-TMLR-2024}, DINOv2 with Registers, and DINOv3 \cite{Siméoni-arXiv-2025}.
\end{itemize}
For architectures without a CLS token (ConvNeXt, MobileViT, Swin Transformer), the average pooling of all patch tokens from the final layer is fed into the classification head.
Table \ref{tbl:exp_1} summarizes the results under the MICO protocol.
The results indicate that self-supervised models generally outperform supervised counterparts.
This result suggests that self-supervised pre-training yields features with higher transferability and is more effective at capturing discriminative cues for spoofing detection across different datasets.
Among the self-supervised models, DINOv2 with Registers demonstrates the most robust and consistent domain generalization. 
While DINOv3 achieves the highest accuracy in specific scenarios (OMI$\rightarrow$C and ICM$\rightarrow$O), its performance suffers a significant drop in the challenging OCM$\rightarrow$I protocol. 
In contrast, DINOv2 with Registers maintains high accuracy across all protocols without severe degradation. 
As indicated by the overall average AUC, DINOv2 with Registers achieves the highest mean performance ($96.25\%$), effectively outperforming both DINOv3 ($95.14\%$) and CLIP ($95.41\%$). 
This consistency confirms its ability to extract robust, generalized features essential for reliable face anti-spoofing, thereby justifying its selection as the core architecture for our baseline.

\begin{table*}[t]
  \centering
  \caption{Experimental results for MICO [\%].
  The best and second-best results among vision-only methods are highlighted in \textbf{bold} and \underline{underline}, respectively.
  VLM-based multimodal methods are listed at the bottom for reference.}
  \label{tbl:results_MICO}
  \resizebox{\linewidth}{!}{
  \begin{tabular}{lcccccccccc} 
    \toprule
    \multirow{2}{*}{Method} &
      \multicolumn{2}{c}{CIO$\rightarrow$M} &
      \multicolumn{2}{c}{OMI$\rightarrow$C} &
      \multicolumn{2}{c}{OCM$\rightarrow$I} &
      \multicolumn{2}{c}{ICM$\rightarrow$O} &
      \multicolumn{2}{c}{Avg.} \\
      \cline{2-11}
     & HTER$\downarrow$ & AUC$\uparrow$ & HTER$\downarrow$ & AUC$\uparrow$ & HTER$\downarrow$ & AUC$\uparrow$ & HTER$\downarrow$ & AUC$\uparrow$ & HTER$\downarrow$ & AUC$\uparrow$ \\
    \midrule
    DRDG \cite{Liu-IJCAI-2021} & 12.43 & 95.81 & 19.05 & 88.79 & 15.56 & 91.79 & 15.63 & 91.75 & 15.67 & 92.04 \\
    ANRL \cite{Liu-ACM-2021} & 10.83 & 96.75 & 17.83 & 89.26 & 16.03 & 91.04 & 15.67 & 91.90 & 15.09 & 92.24 \\
    SSDG-R \cite{Jia-CVPR-2020} & 7.38 & 97.17 & 10.44 & 95.94 & 11.71 & 96.59 & 15.61 & 91.54 & 11.29 & 95.31 \\
    SSAN-R \cite{SSAN} & 6.67 & \underline{98.75} & 10.00 & 96.67 & 8.88 & 96.79 & 13.72 & 93.63 & 9.82 & 96.46 \\
    PatchNet \cite{Wang-CVPR-2022} & 7.10 & 98.46 & 11.33 & 94.58 & 13.40 & 95.67 & 11.82 & 95.07 & 10.91 & 95.95 \\
    TransFAS \cite{Wang-TBIOM-2022} & 7.08 & 96.69 & 9.81 & 96.13 & 10.12 & 95.53 & 15.53 & 91.10 & 10.64 & 94.86 \\
    DiVT-M \cite{Liao-WACV-2023} & \textbf{2.86} & \textbf{99.14} & 8.67 & 96.92 & \textbf{3.71} & \textbf{99.29} & 13.06 & 94.04 & 7.08 & 97.35 \\
    SA-FAS \cite{SA-FAS} & 5.95 & 96.55 & 8.78 & 95.37 & 6.58 & 97.54 & 10.00 & 96.23 & 7.83 & 96.42 \\
    IADG \cite{Zhou-CVPR-2023} & 5.41 & 98.19 & 8.70 & 96.44 & 10.62 & 94.50 & 8.86 & 97.14 & 8.40 & 96.57 \\
    GAC-FAS \cite{Le-CVPR-2024} & \underline{5.00} & 97.56 & \underline{8.20} & 95.16 & \underline{4.29} & \underline{98.87} & 8.60 & 97.16 & \textbf{6.52} & 97.19 \\
    Li \cite{Li-NN-2024} & 12.92 & 94.33 & 9.26 & \underline{96.98} & 10.87 & 95.46 & 15.13 & 91.43 & 12.05 & 94.55 \\
    DiffFAS-V \cite{Ge-ECCV-2024} & \textbf{2.86} & 98.41 & 10.11 & 96.32 & 6.36 & 97.89 & \underline{8.11} & \underline{97.27} & \underline{6.86} & \underline{97.47} \\
    \midrule
    Baseline & 8.86 & 96.95 & \textbf{4.49} & \textbf{98.92} & 9.81 & 96.70 & \textbf{7.35} & \textbf{98.07} & 7.63 & \textbf{97.66} \\
    \midrule
    FLIP \cite{Srivatsan-ICCV-2023} & 4.95 & 98.11 & 0.54 & 99.98 & 4.25 & 99.07 & 2.31 & 99.63 & 3.01 & 99.20 \\
    CFPL \cite{Liu-ICCV-2023} & 1.43 & 99.28 & 2.56 & 99.10 & 5.43 & 98.41 & 2.50 & 99.42 & 2.98 & 99.05 \\
    I-FAS \cite{Zhang-AAAI-2025} & 0.32 & 99.88 & 0.04 & 99.99 & 3.22 & 98.48 & 1.74 & 99.66 & 1.33 & 99.50 \\
    \bottomrule
  \end{tabular}
  }
\end{table*}

\noindent
{\bf (ii) Cross-dataset Evaluation}: 
We perform cross-dataset evaluation using the MICO protocol to validate the domain generalization capability of the baseline method, with results summarized in Table \ref{tbl:results_MICO}. The methods are grouped into conventional vision-only approaches (first group), the baseline itself (second group), and VLM-based SOTA methods (third group). Compared with conventional vision-only methods, ours shows overall competitive performance across most transfer settings, though not uniformly the best. In CIO$\rightarrow$M, it achieves an AUC comparable to several strong image-only approaches and close to the top-performing methods in this group. On OMI$\rightarrow$C, it exceeds the reported results of conventional methods under this protocol. For OCM$\rightarrow$I, however, our baseline has slightly lower performance than several recent approaches. This suggests that while DINOv2 with Registers provides solid cross-domain generalization, there remains room for improvement under this particular domain shift. On ICM$\rightarrow$O, it also exceeds the reported results of conventional methods under this protocol.
Overall, these results indicate that this method offers stable and competitive performance across different cross-dataset scenarios, without consistently dominating every setting. When compared to VLM-based SOTA methods (FLIP, CFPL, and I-FAS), it remains competitive despite its simpler design.
While these large-scale VLM approaches achieve the highest AUC scores (all above 99\% in several settings), the baseline method narrows the gap substantially, for example, 98.92\% vs. 99.99\% on OMI$\rightarrow$C and 98.07\% vs. 99.66\% on ICM$\rightarrow$O, without relying on extremely large multimodal backbones.
Beyond direct performance comparisons, these state-of-the-art VLM-based methods primarily employ the CLIP image encoder for visual feature extraction.
As demonstrated in our benchmarking results (Table \ref{tbl:exp_1}), self-supervised models like DINOv2 with Registers capture more robust and fine-grained visual cues than CLIP, even without relying on semantic alignment.
Therefore, our established vision-only baseline is not only an effective independent approach but also has the potential to serve as a superior visual backbone for these multimodal frameworks.
Replacing their existing image encoders with our optimized feature extractor could further enhance the performance of future VLM-based FAS systems.
Overall, the results indicate that this baseline provides a strong domain-invariant feature representation. Despite not leveraging massive VLM architectures, it achieves highly competitive cross-dataset performance, demonstrating that carefully designed vision-only models can approach the performance of large VLM-based systems while maintaining architectural simplicity and efficiency.

\begin{table}[t]
  \centering
  \caption{Experimental results for limited source domains [\%]. The best and second-best results are highlighted in \textbf{bold} and \underline{underline}, respectively.}
  \label{tbl:LSD}
  \begin{tabular}{@{}lcccc@{}}
    \toprule
    \multirow{2}{*}{Method} & \multicolumn{2}{c}{MI$\rightarrow$C} & \multicolumn{2}{c}{MI$\rightarrow$O} \\
    \cmidrule(lr){2-3} \cmidrule(lr){4-5} & HTER$\downarrow$ & AUC$\uparrow$ & HTER$\downarrow$ & AUC$\uparrow$ \\
    \midrule
    DRDG \cite{Liu-IJCAI-2021} & 31.28 & 71.50 & 33.35 & 69.14\\
    ANRL \cite{Liu-ACM-2021} & 31.06 & 72.12 & 30.73 & 74.10\\
    SSDG-M \cite{Jia-CVPR-2020} & 31.89 & 71.29 & 36.01 & 66.88 \\
    SSAN-R \cite{SSAN} & 30.00 & 76.20 & 29.44 & 76.62\\
    DiVT-M \cite{Liao-WACV-2023} & 20.11 & 86.71 & 23.61 & 85.73\\
    IADG \cite{Zhou-CVPR-2023} & 24.07 & 85.13 & 18.47 & 90.49\\
    GAC-FAS \cite{Le-CVPR-2024} & 16.91 & 88.12 & 17.88 & 89.67\\
    FAS-Aug \cite{Cai-IJCV-2024} & 16.89 & 90.06 & \underline{15.10} & \underline{92.69} \\
    DiffFAS-V \cite{Ge-ECCV-2024} & \underline{15.06} & \underline{92.83} & 16.19 & 92.62\\
    \midrule
    Baseline & \textbf{8.29} & \textbf{97.10} & \textbf{12.11} & \textbf{95.36} \\
    \bottomrule
\end{tabular}
\end{table}

\noindent
{\bf (iii) Limited Source Domain}: 
We further evaluate the robustness under resource constraints using the limited source domain  setting of the MICO protocol.
This challenging configuration uses only two source datasets (MSU-MFSD and Idiap Replay-Attack) for training, testing the model's ability to generalize from severely restricted domain diversity.
Table \ref{tbl:LSD} shows the experimental results.
While prior vision-only methods exhibit a severe performance drop under this setting, our approach maintains exceptional accuracy.
We achieve state-of-the-art results among vision-only methods, surpassing all previous works (e.g., HTER 8.29\% and AUC 97.10\% on MI$\rightarrow$C).
This superiority is obtained despite using a significantly light architecture.
This result strongly validates the effectiveness of our vision-only method in capturing domain-invariant cues and adapting to unseen conditions, a key property for resource-constrained deployment.

\begin{table}[t]
  \centering
  \caption{Parameter count for different methods. The lowest parameter counts are highlighted in \textbf{bold}.}
  \label{tbl:TotalParam}
  \begin{tabular}{@{}lcc@{}}
    \toprule
    Method & Trainable Params [M] & Total Params [M] \\
    \midrule
    CFPL \cite{Liu-ICCV-2023} & 94 & 157 \\
    FLIP \cite{Srivatsan-ICCV-2023} & 170 & 170 \\
    I-FAS \cite{Zhang-AAAI-2025} & 104 & 3,100 \\
    \midrule
    Baseline & \textbf{87} & \textbf{87} \\
    \bottomrule
\end{tabular}
\end{table}


\noindent
{\bf (iv) Parameter Count and Computational Complexity}:
A comparison of model sizes across recent state-of-the-art FAS methods as seen in Table \ref{tbl:TotalParam} clearly illustrates the substantial parameter efficiency of our DINOv2-based approach.
I-FAS \cite{Zhang-AAAI-2025} relies on a large-scale multimodal architecture that integrates a CLIP ViT-L/14 image encoder (approximately 304M parameters) with the OPT-2.7B language model (approximately 2.7B parameters), resulting in a total model size of roughly 3.1B parameters. Although only 104M parameters (the GAC module) are trainable during optimization, the full architecture must still be stored and executed during inference, leading to considerable computational and memory overhead.
CFPL \cite{Liu-ICCV-2023} adopts a more compact multimodal design built upon CLIP ViT-B (approximately 86M parameters) and its 63M-parameter text encoder, augmented with two lightweight Q-Formers (approximately 3.5M parameters each).
This results in a total model size of approximately 157M parameters, with 94M trainable parameters during training.
Similarly, FLIP \cite{Srivatsan-ICCV-2023} employs the same CLIP ViT-B backbone and text encoder, reaching approximately 150M-170M parameters during training.
While FLIP reduces inference complexity by retaining only the 86M-parameter image encoder after precomputing text embeddings, its training process still depends on a multimodal framework.
In contrast, our DINOv2 with Registers method adopts a purely vision-based architecture without any language model or auxiliary text encoder.
Using DINOv2 ViT-B/14 with Registers (approximately 86M parameters), our approach achieves competitive cross-domain generalization performance while maintaining a total parameter count of only 87M.
This corresponds to less than 3\% of the parameters required by I-FAS and roughly 55\% of CFPL's full architecture, while remaining comparable to FLIP's inference-time model size.
Importantly, our results demonstrate that similar generalization performance can be achieved without the substantial parameter overhead inherent to vision-language models.
By relying solely on self-supervised visual representations, our baseline method avoids the architectural complexity, memory footprint, and computational burden associated with multimodal large language models, highlighting that effective FAS generalization does not necessitate a VLM-based design.

\section{Conclusion}

In this paper, we revisited the potential of vision-only foundation models for FAS to address the computational limitations of recent VLMs. 
Through a comprehensive benchmarking of 15 diverse pre-trained vision models under severe cross-domain scenarios, we demonstrated that self-supervised vision models possess superior domain generalization capabilities. 
Based on these insights, we established a robust and highly efficient vision-only baseline utilizing DINOv2 with Registers combined with FAS-Aug, PDA, and APL. 
Extensive experiments demonstrated that our baseline not only achieves competitive performance under the standard MICO protocol but also establishes a new state-of-the-art in data-constrained LSD scenarios.
The results also indicated that an optimized self-supervised vision transformer provides a highly practical standalone solution for resource-constrained applications, while also offering a potent visual backbone to further advance future multimodal FAS systems.

\section{Acknowledgment}

This work was supported in part by JSPS KAKENHI 23H00463 and 25K03131, JST BOOST JPMJBS2421, and the WISE Program for AI Electronics, Tohoku University.

{
    \small
    \bibliographystyle{ieeenat_fullname}
    \bibliography{main}
}


\end{document}